\documentclass[10pt,twocolumn,letterpaper]{article}

\usepackage{customization}

\usepackage{iccv}

\usepackage[pagebackref,breaklinks,colorlinks]{hyperref}

\hypersetup{
    linkcolor=eq-fig-tab-color,  
    citecolor=ref-color,  
}

\usepackage[numbers, sort&compress]{natbib}

\iccvfinalcopy 

\begin{document}

\title{Efficient Controllable Multi-Task Architectures}

\author{Abhishek Aich$^{\star,\dagger}$, Samuel Schulter$^{\dagger}$, Amit K. Roy-Chowdhury$^{\star}$,\\ Manmohan Chandraker$^{\dagger,\ddagger}$, Yumin Suh$^{\dagger}$\\
$^{\star}$University of California, Riverside, $^{\dagger}$NEC Labs America, $^{\ddagger}$University of California, San Diego\\
}

\maketitle
\setlength{\textfloatsep}{0.5cm}
\begin{abstract}
    We aim to train a multi-task model such that users can adjust the desired compute budget and relative importance of task performances after deployment, without retraining. This enables optimizing performance for dynamically varying user needs, without heavy computational overhead to train and save models for various scenarios. To this end, 
    we propose a multi-task model consisting of a shared encoder and task-specific decoders where both encoder and decoder channel widths are slimmable. Our key idea is to control the task importance by varying the capacities of task-specific decoders, while controlling the total computational cost by jointly adjusting the encoder capacity. This improves overall accuracy by allowing a stronger encoder for a given budget, increases control over computational cost, and delivers high-quality slimmed sub-architectures based on user’s constraints. Our training strategy involves a novel `Configuration-Invariant Knowledge Distillation' loss that enforces backbone representations to be invariant under different runtime width configurations to enhance accuracy. Further, we present a simple but effective search algorithm that translates user constraints to runtime width configurations of both the shared encoder and task decoders, for sampling the sub-architectures. The key rule for the search algorithm is  to provide a larger computational budget to the higher preferred task decoder, while searching a shared encoder configuration that enhances the overall MTL performance. 
    Various experiments on three multi-task benchmarks (PASCALContext, NYUDv2, and CIFAR100-MTL) with diverse backbone architectures demonstrate the advantage of our approach. For example, our method shows a higher controllability by $\sim$ 33.5\% in the NYUD-v2 dataset over prior methods, while incurring much less compute cost.
\end{abstract}

\section{Introduction}
Multi-task learning (MTL) often aims to solve multiple related tasks together using a single neural network for economy of deployment \cite{zhang2021survey, vandenhende2021multi}. 
Humans can handle multiple tasks with diverse trade-offs (e.g., due to availability of resources, adaptable reaction time, etc.), however, most existing MTL architectures are incapable of transforming themselves to handle multiple user constraints without being retrained for each scenario. In this paper, we address the problem of designing controllable dynamic convolutional neural networks (CNN) for MTL that can adjust jointly for two types of user requirements, task preference and compute budget.
\begin{figure}[!t]
    \hspace*{-0.5em}
    \centering
    \includegraphics[width=\columnwidth]{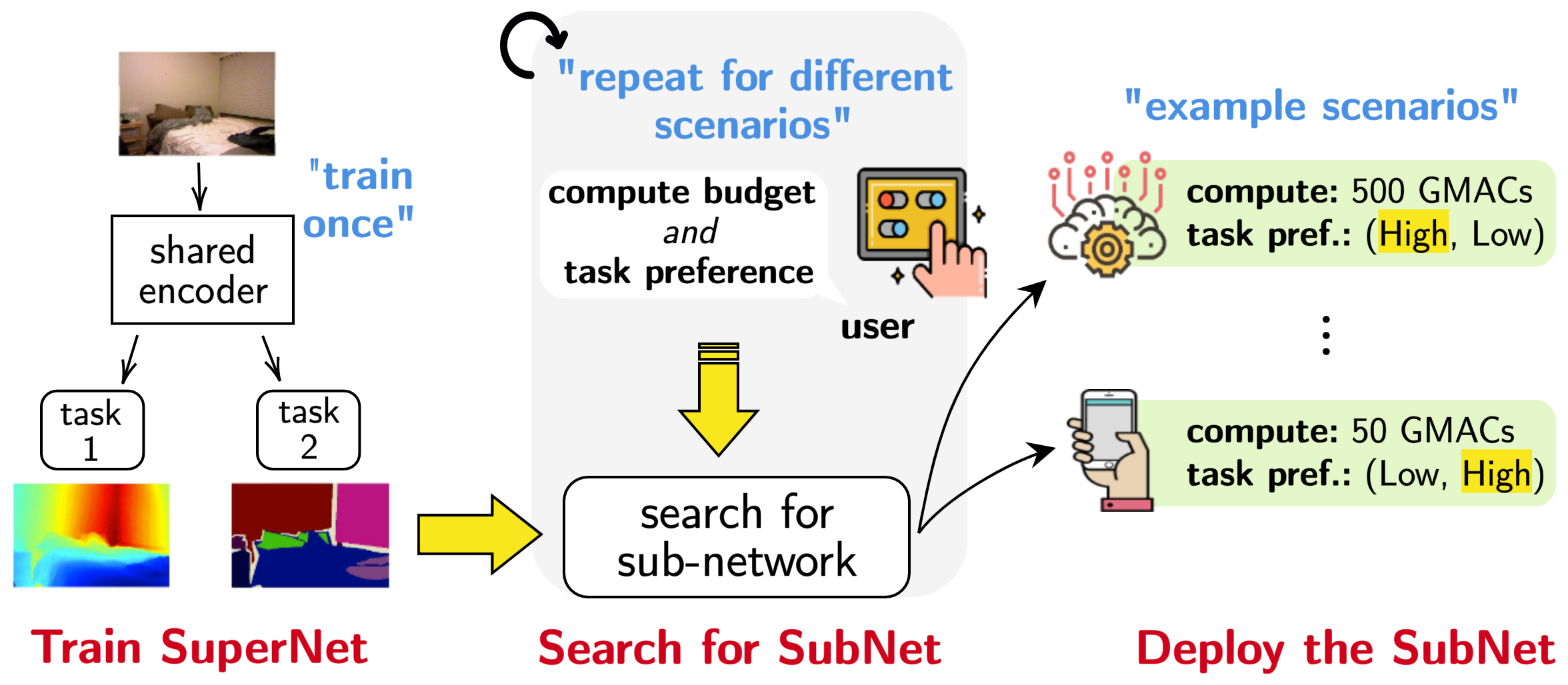}
    \caption{\textbf{Problem Setup.} We aim to provide users precise control on compute allocation as per their MTL performance preference, with the ability to change these dynamically without re-training. To accomplish this, we provide a strategy where a MTL SuperNet is trained only once but allows crafting SubNets that can be sampled based on the user's MTL constraints (compute cost \textit{and} task preference) at test-time. ``High" task preference for task $i$ implies the performance for task $i$ is more important than other tasks.}
    \label{fig:teaser}
    \vspace*{-0.5em}
\end{figure}

 Real-world MTL systems are seeing growing applications ranging from autonomous cars \cite{karpathy2019multi} to video cameras for traffic analysis \cite{chen2019edge}, with respective task performance preferences. \tcomment{For example, observe Fig. \ref{fig:teaser}. A single MTL architecture can allow two users to use the same model but with custom task preferences based on the available compute cost. The user with higher compute (\eg self-driving cars) may expect higher performance on task \textit{1}, but the user with lower compute (\eg traffic cameras) would prefer higher performance on task \textit{2} given the budget.} It will be extremely inefficient to create and train MTL architectures for all such possible variations of user requirements due to expensive design and deployment costs \cite{cai2020once, yu2020bignas, raychaudhuri2022controllable}. This brings forth the need for flexible MTL architectures that allow test-time trade-offs based on relative task importance \textit{and} resource allocation. Some prior methods have introduced  dynamic MTL networks \cite{momma2022multi, lin2020controllable,navon2020learning,mahapatra2020multi} in an effort to incorporate changing user task preferences at test-time. However, such methods do not account for changing the user's computational budget as they assume fixed computation cost resulting in limited applicability. Recently, Controllable Dynamic Multi-task Architecture (CDMA) \cite{raychaudhuri2022controllable} introduced a multi-stream (equal to number of tasks) architecture to handle both changing task preferences and compute budgets. For controllability, it adjusts branching locations in the encoder and generates encoder weights using external hypernetworks~\cite{ha2016hypernetworks} while fixing decoders. 
%
In this paper, we propose a multi-task method called `Efficient Controllable Multi-Task architectures' (\altar) that consists of a shared encoder and task-specific decoders where the channel widths of both modules are slimmable~\cite{yu2018slimmable}. Our key idea is to control the task importance by varying the capacities of task-specific decoders while controlling the compute budget by jointly adjusting the encoder capacity.
This is based on our observation that a larger backbone achieves overall higher multi-task accuracy (even with task conflicts) compared to separately trained multiple smaller backbones. Further, in contrast to adjusting the branching points of multiple encoder streams~\cite{raychaudhuri2022controllable}, our approach can achieve overall higher controllability by adopting one stronger backbone for a given compute budget. 
Since decoder widths affect both accuracy and computational cost by a considerable amount, especially for dense prediction tasks, adjustment of decoder capacities is sufficient to control the task preferences. Constraining to control the task preference only through the decoder capacities further avoids adversarial effects when changing the shared encoder as it may cause different effects to each task, which is hard to control. Finally, adjusting both decoder and encoder largely increases the control over the computational cost.

As the training is performed only once and the encoder is shared among tasks, \altar optimizes the sub-architectures by distilling \cite{hinton2015distilling} the encoder knowledge of the parent architecture, that is capable of handling task conflicts given its large capacity \cite{tang2022you,xin2022current,kurin2022defense}. In particular, it uses a novel `Configuration Invariant knowledge distillation' (CI-KD) strategy to make the embeddings of the shared encoder invariant to the varying sub-architecture configuration. At test-time, \altar uses the joint constraints and extracts a sub-architecture by searching for the most suitable encoder and decoder width configuration using the proposed evolution-based algorithm \cite{real2019regularized} designed for MTL models. The key rule for the search algorithm is to provide a larger computational budget to the higher preferred task decoder, while searching a shared encoder configuration that enhances overall MTL performance. 

\tcomment{Interestingly, without any need for external hypernetworks (to predict large tensor weights of the parent architecture) and with a shared encoder (that allows task scalability), \altar demonstrates strong \textit{task preference} - \textit{task accuracy} - \textit{efficiency} trade-offs. Our extensive experiments on benchmark datasets demonstrate strong MTL controllability across a wide range of joint preferences (\eg, an increase in Hypervolume \cite{zitzler1999multiobjective} of $\sim$34\% is observed when compared to state-of-the-art \cite{raychaudhuri2022controllable} during testing in the NYUD-v2 dataset \cite{silberman2012indoor}).}
To summarize, our contributions in this paper are:
\begin{enumerate}[wide, labelwidth=!, labelindent=0pt, itemsep=0pt]
    \item We present a new method to sample high-performing efficient MTL sub-architectures from a single MTL SuperNet that can satisfy both user preferences of task performance and computational budget, dynamically without retraining.
    \item Our method includes two key components:
    \begin{enumerate}[wide, labelwidth=!, labelindent=0pt, itemsep=0pt, topsep=0pt]
        \item[$\bullet$] A \textit{training} strategy to enhance the MTL performance of sub-architectures in order to have minimal performance drop even if user's constraints become restricted. In particular, it uses a CI-KD loss to transfer the encoder knowledge of the parent model, which is capable of handling multi-task conflicts, to the encoders of sub-models.  

        \item[$\bullet$] A subsequent \textit{search} strategy that translates the task preferences to sample the task decoders for better performance and searches for shared encoder width configuration that supports the decoders for overall better MTL performance.
    \end{enumerate}
    \item We show superior controllability on sampling sub-models compared to prior methods. For example, we show a higher controllability by $\sim$ 33.5\% in the NYU-v2 \cite{silberman2012indoor} (3 tasks) dataset and $\sim$ 55\% in Pascal-Context \cite{mottaghi2014role} (5 tasks) dataset over state-of-the-art method CDMA \cite{raychaudhuri2022controllable}.
\end{enumerate}

\section{Related Works}
\label{sec:related_works}
\paragraph{Multi-task Learning (MTL).} Growing demands for MTL capable systems has led to a huge growth in methods for designing effectual architectures that can leverage \textit{shared} feature representations \cite{ruder2017overview,zhang2021survey,vandenhende2021multi} among tasks. Broadly, either these methods (\textit{a}) \cite{misra2016cross,gao2019nddr,ruder2019latent,yang2016trace,yang2016deep,rusu2016progressive,long2015learning} provide \tcomment{individual encoders (with respective decoders) for each task} and optimize to decrease the distance between the parameters (\textit{soft} sharing), or (\textit{b}) optimize a single (shared) encoder for all tasks  \cite{bhattacharjee2022mult,kendall2018multi,lin2019pareto,mahapatra2020multi,caruana1998multitask,teichmann2018multinet,kokkinos2017ubernet} followed by individual decoders  (\textit{hard} sharing). A significant number of prior works mainly focus on building MTL strategies under static (or fixed) task preference and compute budget. Hence, any change in these constraints results in re-training the model from scratch. In order to save on designing costs and stringently follow user requirements, some prior works have presented strategies to obtain custom MTL architectures by search mechanisms \cite{gao2020mtl,vu2023toward,bruggemann2021exploring,guo2020learning,sun2019adashare,vandenhende2019branched,standley2020tasks,bruggemann2020automated} similar to traditional neural architecture search (NAS). Such MTL NAS methods are designed for a single model matching only \textit{one} compute budget and are unsuitable for dynamic changing requirements. Different from these, we aim to provide a novel solution for MTL under dynamic (or changing) task preference and compute budget without re-training. 
\paragraph{Controllable MTL Neural Networks.} 
\cite{raychaudhuri2022controllable,mahapatra2020multi,lin2019pareto,lin2020controllable,navon2021learning} have proposed controllable MTL architectures where a parent network is trained to extract sub-architectures that follows user's constraints without re-training. For example, \cite{lin2020controllable,navon2021learning} presented strategies to train hypernetworks to predict weights of a MTL SuperNet in order to match user's \textit{changing} task performance preference. A major drawback of these approaches is that they assume a static compute budget, making them incapable of handling dynamic compute resources. Some prior single task learning (STL) methods such as Once-for-all networks \cite{cai2020once} and BigNAS \cite{yu2020bignas} provide train and search strategies for dynamic compute resources. However, by design they do not provide any recipe for incorporating user MTL task preferences. In short, none of these aforementioned works provide all of dynamic and multi-task aspect as ECMT. Different from \cite{lin2020controllable,navon2021learning,lin2019pareto}, CDMA \cite{raychaudhuri2022controllable} presented a method to handle multiple or joint MTL user constraints where both task preference and compute budget constraints are dynamic. It uses two hypernetworks to predict both network architecture and parameters to match joint constraints. Further, its parent model follows the soft-parameter sharing MTL setup. As a result, CDMA requires additional memory overhead due to two external hypernetworks, faces scalability issues due to soft sharing of encoder, and doesn't consider the decoder in it's search space. Better than CDMA, \altar uses a novel training paradigm for hard-parameter sharing MTL architecture and uses a simple strategy to sample high performing sub-architectures. The search space is defined by the layer width \cite{yu2019universally, yu2018slimmable} and encompasses both the shared encoder and the task decoders. 
\begin{figure*}[!ht]
    \centering
    \includegraphics[width=\textwidth]{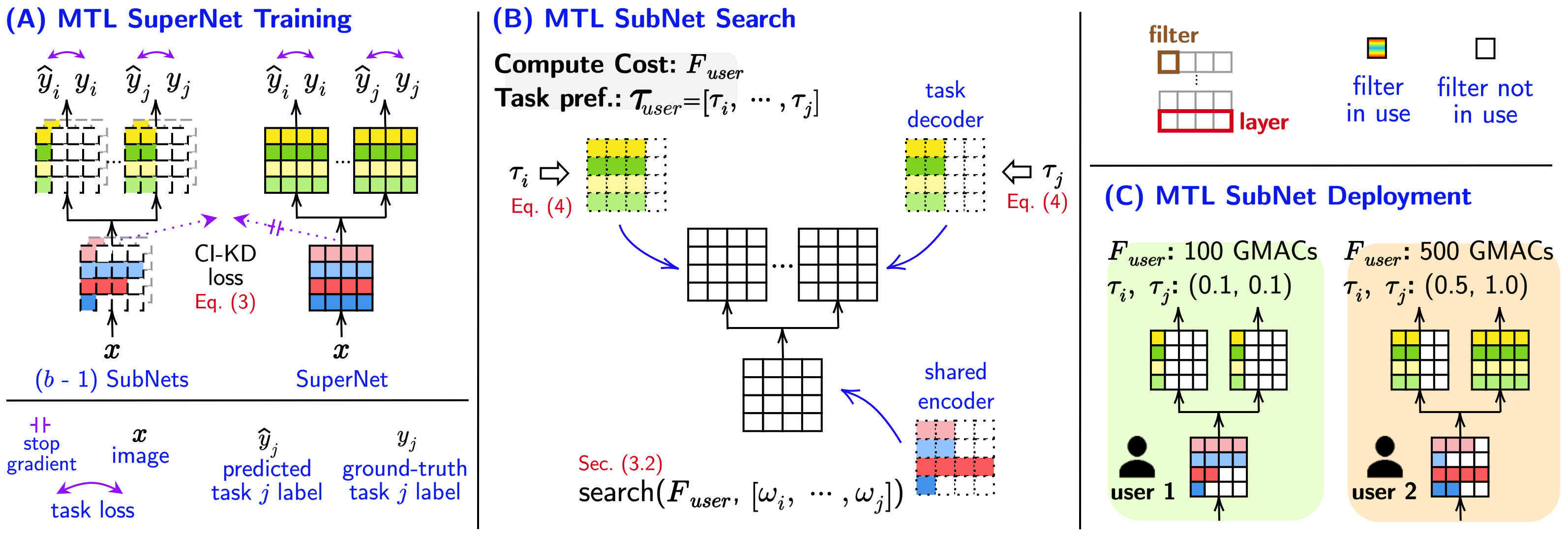}
    \caption{\textbf{Illustration of our pipeline.} Our overall pipeline to obtain MTL SubNets is presented. In (\textbf{A}), we present the SuperNet training strategy where we train the MTL SuperNet collaboratively with the MTL SubNets. We propose a knowledge distillation loss to transfer the knowledge of the largest capacity encoder, which has less task conflicts, to smaller capacity encoders of the SubNets. In (\textbf{B}), we demonstrate our novel strategy to search SubNets (both the shared encoder and decoders) based on the user's joint constraints (task preference and available compute budget). In (\textbf{C}), the search algorithm can provide SubNets as per multiple user preferences.}

    \label{fig:main_fig}
\end{figure*}
\paragraph{Slimmable Neural Networks.} Slimmable networks \cite{yu2018slimmable,yu2019universally,yu2019autoslim} proposed a method to train STL neural networks to support operating using different filter levels controlled by width (or number of filters) multipliers; e.g., a width multiplier
0.5 `slims' down all the parent model layers resulting in a sub-network that has 50\% of filters. These width multipliers are uniform across the architecture, and can be tuned to match the user's memory budget with minimal performance degradation. This strategy has inspired solutions for various STL research areas such as generative adversarial networks \cite{hou2021slimmable}, semantic segmentation \cite{xue2022slimseg} and others \cite{li2021dynamic,chin2021joslim}. Similar to \cite{yu2018slimmable}, \cite{cai2020once, yu2020bignas} explored different freedom directions such as depth and kernel sizes to achieve strong accuracy-efficiency trade-off in STL setup. While we leverage the idea of slimmable models, ECMT is strikingly different from the aforementioned works as they do not support the capability of task performance preference trade-off. Further, \cite{li2021dynamic,chin2021joslim, cai2020once, yu2020bignas} have been mostly explored only for classification tasks. ECMT is designed for MTL setup that extends beyond classification tasks to dense prediction tasks.

\section{Problem Statement}
\label{sec:problem_statement}

\paragraph{Notations.} We denote data distribution as composed of training set $\D_{tr}$, validation set $\D_{val}$, and testing set $\D_{te}$ with $N$ tasks. Each task shares the input image $\x$ with corresponding outputs $\mathcal{Y} = $ \{\y{1}, \y{2}, $\cdots$, \y{N}\}. We create a MTL parent architecture or SuperNet $\Snet$ composed of a single shared encoder among tasks and $N$ task decoders. $\Snet$ is end-to-end non-uniformly slimmable: every layer can be tuned to have it's own set of filters $\mathcal{F}$ (also called \textit{width} \cite{yu2018slimmable}) which is independently controlled by a separate width ratio $\omega \in (0, 1]$. Let $\WList = [\omega_{\min}, \cdots, \omega_{\max}]$ be the set of possible values of width ratios with $\omega_{\max}$ ($\omega_{\min}$) representing the maximum (minimum) possible values. A sub-network or SubNet $\SnetC$ can be created from $\Snet$ by setting width ratios for all $L$ layer denoted using a $L$-tuple $\SnetC = \langle\omega^{(\zeta)}_{1}, \omega^{(\zeta)}_{2}, \cdots, \omega^{(\zeta)}_{L}\rangle\in\WList\times\WList\times\cdots\WList$. Let the set of loss functions for the $N$ tasks be $\{\Loss{1}, \Loss{2}, \cdots, \Loss{N}\}$. Formally, task preference indicates amount of available compute budget for the given task. We denote the task preference list be $\TList = [\tau_1, \tau_2, \cdots, \tau_N]$ with $\tau_i \in [0.0, 1.0]$, where higher value indicates higher preference. Finally, we denote the weights of $\Snet$ as $\bm{\theta}$. 
\paragraph{Problem Statement.} 
Our goal is to train the MTL SuperNet $\Snet$ that allows crafting multiple MTL SubNets $\SnetC$ operable for a wide range of joint MTL user budgets (compute budget $F_{user}$, and task preference $\TList_{user}$) with minimal performance drop. The SuperNet $\Snet$ (and the SubNets $\SnetC$) takes image $\x$ as input and predicts $N$ task outputs $\mathcal{Y}$. To train $\Snet$, we define the following problem:
\begin{align}
    \argmin_{\bm{\theta}}~~\underset{\x, \mathcal{Y} \sim\D_{tr}}{\mathbb{E}} 
    \sum_{n=1}^N \rho_n\Loss{n}\big{(}\bm{\theta}\big{)}
     \label{eq:train}
\end{align}
Here, $\rho_n$ is the weight of $n^{th}$ task loss. Once $\Snet$ is trained, the joint constrained search for obtaining $\SnetC$ can be expressed as the following problem:
\begin{gather}
\min_{\SnetC}~~\underset{\x, \mathcal{Y} \sim\D_{val}}{\mathbb{E}}
\sum_{n=1}^N \rho_n\Loss{n}\big{(}\SnetC\big{)}\notag\\
\label{eq:search}
\begin{aligned}
\text{s.t. } 
\text{compute}(\SnetC)\leq F_{user}, \\
\text{task preference}(\SnetC) = \TList_{user} 
\end{aligned}
\end{gather} 

\section{Proposed Method}
\label{sec:method}

During training, we solve the problem in Eq. \ref{eq:train} by constructing a SuperNet $\Snet$ parameterized by layer-wise width ratios in $\WList$ (see Sec. \ref{sec:train}). During inference, we solve Eq. \ref{eq:search} \tcomment{by searching for the most suitable encoder and decoder width configuration} using an evolution-based search algorithm based on the joint constraints (see Sec. \ref{sec:search}). The training is performed only once, whereas the search is performed for each deployment scenario. 

\subsection{Training the MTL SuperNet}
\label{sec:train}
\begin{figure}[!t]
    \hspace*{-0.8em}
    \centering
    \includegraphics[width=\columnwidth]{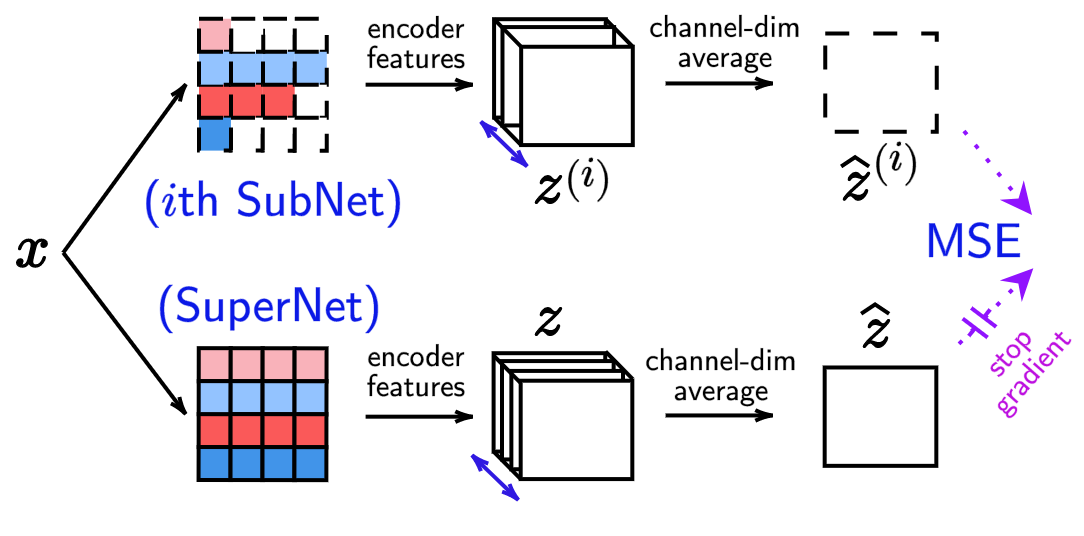}
    \caption{\textbf{Illustration of our Configuration Invariant Knowledge Distillation (CI-KD) Loss.} This loss encourages the shared encoder features to be invariant of the $i^{th}$ SubNet capacity, by enforcing them to be close to the SuperNet's shared encoder features.}
    \label{fig:kd_loss}
\end{figure}

\paragraph{Overview.} Fig. \ref{fig:main_fig}-(\textbf{A}) provides an overview of the proposed training pipeline. At its core, the training strategy aims to enhance the MTL performance of the SubNets sampled from the SuperNet in order to have minimal performance drop even if user's constraints become restricted. To this end, we leverage the Sandwich Rule (SR) training \cite{yu2019universally} and make necessary modifications for our MTL setup. In particular, it involves a novel (width) configuration-invariant knowledge distillation loss aimed at teaching the SubNet encoders from the SuperNet encoder. We now discuss the pipeline in detail. 
\paragraph{Training with the Sandwich Rule.} The SR training \cite{yu2019universally} for single-task learning (STL) requires that in each training iteration, we update the SuperNet with the collectively accumulated loss gradients of the model at $b$ widths. We follow \cite{yu2019universally} and choose $b=4$ which includes the model at largest width $\omega_{\max}$, smallest width $\omega_{\min}$, and $b-2$ randomly chosen models at non-uniform widths. Further, the STL SubNets in SR \cite{yu2019universally} are optimized \textit{only} using the predictions of the largest width model (\ie SuperNet). We build upon this rule and introduce the following changes. \\
\indent In contrast to the aforementioned, we enforce each SubNet to learn the MTL data distribution directly from the available ground-truth labels $\bm{y}$. The training loss of collective learning (\ie training each SubNet as the SuperNet from ground-truth labels) is denoted as $\Lco$. Training SubNets with ground-truth labels helps us avoid the following pitfall: we do not need to train the MTL SubNets from the output predictions of a weak parent MTL model (i.e. predictions in the initial iterations can be weak as it is being trained from scratch). We could tackle this by training the parent model standalone, but this is contrary to our goal of saving on design costs of training \textit{once}. Furthermore, we could also adopt knowledge distillation strategies proposed in prior MTL works \cite{li2020knowledge, jacob2023online}. However, doing so would create additional training overhead for STL models for each task, which is also not our end-goal. Hence, we present a new methodology of distilling the knowledge of the parent model $\Snet$ to SubNets without using output predictions, which brings us to our proposed encoder-based knowledge distillation (KD) loss.
\paragraph{Configuration Invariant KD (CI-KD) Loss.} Our CI-KD loss is an \textit{in-place distillation loss} $\Lkd$, which transfers the encoder knowledge of $\Snet$ to the encoders of sub-networks $\SnetC$. The encoder of $\Snet$ is capable of handling multi-task conflicts due to its high capacity \cite{tang2022you,xin2022current,kurin2022defense} and we aim to teach the encoders of the smaller models from its features. In particular, we propose to minimize distance between the encoder features  computed from parent model $\Snet$ and all the $i$th child model involved in the sandwich setup. Now, this loss cannot be directly estimated: the features of $\Snet$'s encoder $\z$ and other child models $\z^{(i)}$ are of different sizes due to the different configurations of the SubNet encoders. To make the shared encoder feature size \textit{configuration invariant}, we compute the average features along the channel dimensions for all models in the Sandwich. We then minimize the mean square error loss between these channel-averaged features of the parent model and the $b-1$ child models as follows. 
\begin{align}
    \Lkd = \dfrac{1}{b-1}\sum_{i}^{b-1} \text{~MSE}(\wz, \wz^{(i)})
    \label{eq:kd_loss}
\end{align}
Here, $\wz$ and $\wz^{(i)}$ are the encoder features of $\Snet$ and $i$th child model (in the Sandwich), averaged along the channel dimension. Note that $\wz$ is detached from the computational graph as we do not intend to update the parent model with $\Lkd$. This distillation loss has been illustrated in Fig. \ref{fig:kd_loss}. To summarize, the SuperNet learning loss only includes Eq. \ref{eq:train}, whereas the $i^{th}$ SubNet learning loss includes both with Eq. \ref{eq:train} and $\lambda\Lkd$ ($\lambda$ is the weight of the CI-KD loss). The training algorithm is given in Algo. \ref{algos:supernet_train_algo}.

{
\begin{algorithm}[!t]
\captionsetup{labelfont={sc,bf}, labelsep=newline, font=small}
\caption{ECMT training for SuperNet \(\Snet\).}
    \begin{algorithmic}[1]
    \Require{Width list $\WList$, ~\eg { \([0.3, 0.4, \cdots 1.0]\)}.}
    \Require{Number of models $b$ in Sandwich Rule,~\eg, \(b=4\).}
    \For {\(t = 1, \cdots, T_{epoch}\)}
        \State{Get input image \(\x\) and task labels \(\mathcal{Y}\).}
        \State{Clear gradients, \texttt{optimizer.zero\_grad()}.}
        \State{Get task outputs \& enc. feature $\z$ from $\Snet$; $\widehat{\mathcal{Y}}, \z = \Snet(\x)$.}
        \State{Get total loss $\Lco$ using \{$\widehat{\mathcal{Y}}, \mathcal{Y}$\} in Eq. \ref{eq:train}.}
        \State{Accumulate gradients, $\Lco$.\texttt{backward()}.}
        \State{Stop gradients of $\z$, \(\z = \z.\texttt{detach()}\).}
        \State{Randomly sample (\(b-1\)) models \{$S_b =\Snet^{(1)}, \cdots,\Snet^{(b-1)}$\}.}
        \State{Add smallest width model to $S_b$.}
        \For {$\Snet^{(i)}$ in $S_b$}
            \State{Get task outputs \& enc. feature $\z$ from $\Snet^{(i)}$}
            \State{Compute loss $\mathcal{L}^{(i)} = \Lco + \lambda\Lkd$.}
            \State{Accumulate gradients, \(\mathcal{L}^{(i)}.\texttt{backward()}\).}
        \EndFor
        \State{Update SuperNet weights, \(\texttt{optimizer.step()}\).}
    \EndFor
    \end{algorithmic}
\label{algos:supernet_train_algo}
\end{algorithm}
}

\subsection{Searching based on Joint User Preferences} 
\label{sec:search}
\paragraph{Overview.} Fig. \ref{fig:main_fig}-(\textbf{B}) provides a simple method to sample sub-networks that follow the user's joint constraints of task preference $\TList_{user}$ and compute budget $F_{user}$. We divide the search into two parts. In \textit{Step 1}, we sample the width ratios of the task decoders based on the user's task preference using a simple rule proposed in Eq. \ref{eq:dec_rule}. With the decoder configuration fixed, we search for an optimal shared encoder configuration in \textit{Step 2}. The goal of the search is to support the sampled task decoders for better performance than randomly choosing the configuration while satisfying the overall compute budget. \textit{Step 2} particularly involves a MTL accuracy predictor that provides quick feedback on the overall model configuration during the search cycle and saves repeated evaluation related computations. We now expand on each step in detail. 

\paragraph{\textit{Step 1}: Setting the task decoders.} We propose to set the width ratios of the task decoders based on the task preference $\TList_{user}=\{\tau_i\}_{i}^{N}$ as they are independent for each task. In particular, we map each $\tau_i$ to the discrete uniform range of $\WList\sim\mathcal{U}(\omega_{\min}, \omega_{\max})$. Assuming $\TList\sim\mathcal{U}(0, 1)$ as a uniform distribution with unit density when $0\leq \tau \leq 1$ (0 otherwise), $\tau_i$ is mapped to a decoder width ratio $\omega_i$ as:
\begin{align}
    \omega_i = \omega_{\min} + (\omega_{\max}-\omega_{\min})\tau_i
    \label{eq:dec_rule}
\end{align}
Clearly, $\omega_i\propto\tau_i$ \ie the decoder of the task with higher preference will be assigned a higher width ratio. This design choice is motivated by the reason to allow a larger computational budget in the available user's budget to the higher preferred task decoder. Once all the decoders are fixed using Eq. \ref{eq:dec_rule}, we search for a shared encoder width configuration to support the aforementioned width decoder configuration. 

\paragraph{\textit{Step 2}: Searching for the shared encoder.} The aim is to sample a width ratio configuration for the shared encoder, that supports the best performance out of the sampled decoders. Randomly choosing the encoder configuration is one possible option, but it doesn't necessarily result in the best architecture due to our large search space. In order to leverage the large search space designed by our non-uniform layer setup and find a better performing model, we use an evolution-based search algorithm \cite{real2019regularized}. The search algorithm consists of three key components. First, we initialize a pool $\mathcal{P}$ of $P$ models ($\mathcal{P}= \{\mathcal{S}^{(1)}, \mathcal{S}^{(2)}, \cdots, \mathcal{S}^{(P)}\}$), all with the \tcomment{fixed} decoder configuration obtained from \textit{Step 1}. Each of these models are characterized by the same width ratio across all encoder layers. Next, we evolve $\mathcal{P}$ in order to find a better performing model than the initialized ones by leveraging the flexibility of choosing width ratios for each layer mutually exclusively in $\SnetC$. We randomly choose $K<L$ encoder layers and change the width ratio $\omega_k$ by the rule:
\begin{align}
    \widehat{\omega}_k = \omega_k + \eta \text{ sign}(F_{user} - F^{(\zeta)})
    \label{eq:search_rule}
\end{align}
Here, $F^{(\zeta)}$ is the computational cost of $\SnetC$ (\eg GMACs), $F_{user}$ is the computational budget set by the user, and sign($\cdot$) extracts the sign of the input variable.We set $\eta=0.1$ as we use the design specification $\omega_i - \omega_j = 0.1$. The motivation of this design choice is to push the shared encoder configuration towards larger capacity, which are known to handle task conflicts better \cite{tang2022you,xin2022current,kurin2022defense}. This evolution step creates a new model $\widehat{\SnetC}$ which is added back to $\mathcal{P}$. In the end, the best performing model from the search is provided for deployment. At all steps, we ensure that each model $\mathcal{P}$ satisfies the user's compute budget constraint. In order to \tcomment{quickly evaluate the quality of models in $\mathcal{P}$}, we build a subsidiary neural network $\mathcal{R}$ that provides a feedback on $\widehat{\SnetC}$'s approximate MTL performance. This MTL accuracy predictor $\mathcal{R}$ eliminates the need for repeated cost of getting the \textit{measured} accuracy by providing a \textit{predicted} accuracy. Specifically, $\mathcal{R}$ is optimized to take $\SnetC$'s width configuration as input and predict the approximate performance of this configuration \cite{cai2020once}. To train $\mathcal{R}$, we first create $K$ examples of [$\Arch$, ($\Loss{1}, \cdots, \Loss{N}$)] pairs by randomly sampling $M$ SubNets with different configurations $\Arch$, and computing their task losses on $\D_{val}$. $\Arch$ contains the list of width ratios computed for the shared encoder and the task decoders. \tcomment{In our experiments, we choose $M = 2000$}. The architecture of $\mathcal{R}$ has been provided in Algo. \ref{algo:mlp_predictor} and training procedure is provided in Sec. \ref{sec:exp}. Note that, we only need to collect the data pairs and train $\mathcal{R}$ once, making this overhead negligible in comparison to training $\Snet$. Further, this cost remains constant regardless of changing user requirements. Fig. \ref{fig:main_fig}-(\textbf{C}) provides some examples of the resultant MTL architectures.
\begin{algorithm}[!t]
\captionsetup{labelfont={sc,bf}, labelsep=newline, font=small}
\caption{ECMT SubNet search.}
    \begin{algorithmic}[1]
    \Require{MTL Accuracy Predictor $\mathcal{R}$, Number of search cycles $T_{cycle}$}
    \Require{Model pool size $P$}
    \Require{User MTL constraints ($F_{user}, \TList_{user}$)}
    \State{Compute and fix decoder widths using $\TList_{user}$ using Eq. \ref{eq:dec_rule}.}
    \State{Initialize the model pool $\mathcal{P}$ with $P$ random encoder widths, including uniform encoder widths, that satisfy memory constraint $F_{user}$ }
    \State{Get the most preferred task $\mathcal{T}$ = $\max(\TList_{user})$}
    \For {\(t = 1, \cdots, T_{cycle}\)}
        \State{Randomly choose a model $\Snet^{(i)}$ from $\mathcal{P}$}
        \State{Change the width of a randomly chosen encoder layer with Eq. \ref{eq:search_rule} in $\Snet^{(i)}$ and create $\widehat{\SnetC}$}
        \State{Add the mutated model $\widehat{\SnetC}$ to $\mathcal{P}$}
        \State{Delete the worst model (which results in highest loss in the preferred task $\mathcal{T}$) computed using $\mathcal{R}$.}
    \EndFor
    \State{Return the width configuration of the best performing model in $\mathcal{P}$ predicted using $\mathcal{R}$.}
    \end{algorithmic}
\label{algos:subnet_search_algo}
\end{algorithm}
We summarize our proposed MTL search algorithm in Algo. \ref{algos:subnet_search_algo}. In Algo. \ref{algos:subnet_search_algo}, we call a candidate model ``mutated" (in \textit{Line} 7) when we change a random encoder layer (in \textit{Line} 6).

\section{Experiments}
\label{sec:exp}
In this section, we demonstrate the ability of \altar to extract  efficient architectures based on a user's joint multi-task learning preferences. We show that our proposed framework can scale to a large number of tasks while allowing proficient \textit{task preference} - \textit{task accuracy} - \textit{efficiency} trade-off with a single MTL model. Through extensive experiments on three datasets with various architectures, we provide new insights to the community regarding controllable MTL architectures which can be re-used for diverse deployment scenarios without re-training. 
\paragraph{Training implementation details.} We use PyTorch \cite{paszke2019pytorch} for our implementation.  

\begin{enumerate}[wide, labelwidth=!, labelindent=0pt, topsep=0pt]
\setlength{\itemsep}{0pt}
    \item[$\bullet$] \textbf{NYUD-v2 and Pascal-Context.} We follow \cite{vandenhende2021multi} to set the following hyperparameters. The SuperNet is trained with Adam Optimizer \cite{kingma2014adam} with learning rate $0.0001$, and weight decay $0.0001$. We use a poly learning rate scheduler. We set the total epochs $T_{epochs}$ as 100 for PASCAL-Context/NYUD-v2. The training batch size is set to 8. The task loss weights $\rho_n$ are set as follows- \textit{NYU-v2} $\rightarrow$ Semantic segmentation: 1, Depth: 1, and Surface normals: 10; \textit{PASCAL-Context} $\rightarrow$ Semantic segmentation: 1, Human parts segmentation: 2, Saliency: 5, Edge: 50, Surface normals: 10. The training data augmentations are set from \cite{vandenhende2019branched}. We further add Random Erasing \cite{zhong2020random} with probability 0.5 to the aforementioned. The shared encoder is initialized with ImageNet \cite{deng2009imagenet} pre-trained weights, whereas the decoders are trained from scratch.
    The CI-KD loss weight $\lambda$ is set to 1 and 0.001 for NYUD-v2 and Pascal-Context, respectively. $\WList$ is set as $[0.6, 0.7, \cdots, 1.0]$ for NYUD-v2, and $[0.5, 0.6, \cdots, 1.0]$ for Pascal-Context. The input channels to the DeepLabv3 decoder for NYUD-v2 and Pascal-Context are set to 256 and 128, respectively.
    
    \item[$\bullet$] \textbf{CIFAR100.} All the task loss weights are set to 1. The batch size is set to 256. We set the total epochs $T_{epochs}$ is set as 75. The SuperNet is trained with Adam Optimizer \cite{kingma2014adam} with learning rate $0.001$, and weight decay $0.0001$. The whole SuperNet is trained from scratch. The CI-KD loss weight $\lambda$ is set to 0.01. $\WList$ is set as $[0.2, 0.3, \cdots, 1.0]$.
\end{enumerate}
All experiments use 90\% of training set for all datasets. All other hyperparameters are set as per default PyTorch settings. 
\paragraph{Evaluation criteria.} Following \cite{raychaudhuri2022controllable}, we use hypervolume (HV) \cite{zitzler1999multiobjective} to measure controllability with respect to task preferences. HV helps to evaluate the preference-loss trade-off curve (or Pareto-front \cite{mahapatra2020multi}) among the different tasks in the loss space \cite{fleischer2003measure}. Basically, HV accounts for both the quality of the models (volume they dominate), and also the diversity, measured in the overlap between dominated regions. Higher value indicates better controllability assuming same memory budget. Bold values indicate best results.
\paragraph{Inference implementation details.} For comparisons with CDMA and PHN for controllability, we follow CDMA's method of task preference sampling. In particular, we sample task preferences $\tau \sim P_\tau$, where $P_\tau$ is a Dirichlet distribution of order $N$ with parameters $[\alpha_1, \alpha_2, \cdots, \alpha_N]$ ($\alpha_i >$ 0). Following \cite{raychaudhuri2022controllable}, we set $\alpha_i = 0.2~\forall~i$ for PASCAL-Context/NYUD-v2, and $\alpha_i = 1.0~\forall~i$ for CIFAR100. HV is computed across 20 task preferences following \cite{raychaudhuri2022controllable}. DeepSpeed \cite{deepspeed} is used to compute the architecture GMACs. For any $\TList_{user}$, we always aim to set the highest width for \textit{most preferred task} for best results. Hence, we min-max norm $\TList_{user}$ before using Eq. \ref{eq:dec_rule}. Any value $\in[0,1]$ can be used for each task preference (sum need not be 1). We set the number of search cycles $T_{cycle}$ as 150 for CIFAR100 and 10 for NYUD-v2/Pascal-Context. We set the model population size to 50. The search runtime is proportional to $T_{cycle}$.

\paragraph{Datasets.} We conduct our experiments using three multi-task datasets, namely PASCAL-Context \cite{mottaghi2014role} (5 tasks) and NYU-v2 \cite{silberman2012indoor} (3 tasks), and CIFAR-100 \cite{krizhevsky2009learning, rosenbaum2017routing} (20 tasks). The PASCAL-Context dataset provides joint semantic segmentation, human parts segmentation, saliency estimation, surface normal estimation, and edge detection. The NYU-v2 dataset provides joint semantic segmentation, depth estimation and surface normal estimation. The CIFAR-100 (MTL) dataset provides a split of 20 five-way classification tasks extracted from the original dataset. Semantic segmentation, saliency estimation
and human parts segmentation are evaluated using mean intersection over
union (mIoU). Depth estimation and surface normals are evaluated using the root mean square error (rmse).
\begin{algorithm}[!t]

\captionsetup{labelfont={sc,bf}, labelsep=newline, font=small}
\caption{Pseudocode of our MTL Accuracy Predictor $\mathcal{R}$.
}
\label{algo:mlp_predictor}
\verbatimfont{\footnotesize}%
\begin{verbatim}
import torch.nn as nn

# ----- multi-task performance predictor
class AccuracyPredictor(nn.Module):
    def __init__(self, input_channels: int,
                       task_names: list):
        super().__init__()
        self.model = nn.Sequential(
            nn.Linear(input_channels, 100),
            nn.ReLU(),
            nn.Linear(100, 100),
            nn.ReLU(),
            nn.Linear(100, 50),
            nn.ReLU()
        )
        
        self.task_names = task_names
        
        self.heads = nn.ModuleDict({task: 
                  nn.Linear(50, 1) for task in 
                  self.task_names})
        
    def forward(self, widths: torch.Tensor):
        feats = self.model(widths)
        preds = {t_id: self.heads[t_id](feats)
              for t_id in self.task_names}
        
        return preds 
\end{verbatim}
\end{algorithm}
\paragraph{Baselines.} As reported in Sec. \ref{sec:related_works}, CDMA \cite{raychaudhuri2022controllable} is our main baseline which designs MTL architectures, jointly controllable with task preference and memory constraint. Further, we compare with PHN \cite{navon2021learning} (and its variation PHN-BN \cite{raychaudhuri2022controllable, navon2021learning}), which \textbf{only} allow task controllability. PHN uses a hypernetwork to predict the weights of a shared backbone based SuperNet conditioned on a task preference. Further, we also compare with MTL-Static models trained for highest user constraints. It provides the upper bound of the HV \altar can achieve. The MTL-Static models and \altar are trained with 90\% samples of the original training split. The other 10\% split is used to train the MTL accuracy predictor. 
\paragraph{Architectures.} Our SuperNet $\Snet$ consists of a shared encoder (backbone) with $N$ task decoders. For dense prediction tasks (PascalContext, NYUD-v2), we use the DeepLabv3 architecture \cite{chen2017rethinking} for each task, whereas we use linear layers for classification tasks (CIFAR100-MTL). We follow CDMA and choose the backbone architectures as MobileNetV2 \cite{sandler2018mobilenetv2} for PASCAL-Context, ResNet34 \cite{he2016deep} for NYUD-v2, and ResNet9 \cite{res9} for CIFAR100-MTL. MobileNetV2 and ResNet34 are pre-trained on ImageNet \cite{deng2009imagenet}. Note that, in all cases, both the encoder and the task decoders are non-uniformly slimmable. Following \cite{yu2019universally, yu2020bignas}, we do not accumulate batch norm statistics while training the SuperNet as these are ill-defined due to varying SubNet configurations. During testing, we re-calibrate the batch norm statistics for each SubNet without any fine-tuning of its parameters. 
\paragraph{MTL accuracy predictor details.} The MTL accuracy predictor $\mathcal{R}$ is a three layer feed-forward neural network summarized in Algo. \ref{algo:mlp_predictor}. Given a SubNet model $\SnetC$, we create a list  $\Arch$ consisting of width ratios starting with the shared encoder, followed by the decoder's width ratios. $\Arch$ is then fed to the $\mathcal{R}$ to get the predicted MTL accuracy for all $N$ tasks. To train $\mathcal{R}$, we first create $M=2000$ examples of [$\Arch$, ($\Loss{1}, \cdots, \Loss{N}$)] pairs by randomly sampling $M$ SubNets with different configurations $\Arch$, and computing their task losses on $\D_{val}$. We train this predictor for 150 epochs using L1-loss using the Adam optimizer with a learning rate of $0.001$. The batch size is set to 16.
\begin{table}[t]
\caption{\textbf{Evaluation on NYUD-v2}. Reference point for HV: $[4, 4, 4]$. $\dagger$ indicates smaller compute cost (GMACs).}
\vskip0.5em
\centering
\resizebox{\columnwidth}{!}{
\begin{tabular}{lccccc}
\toprule
\textbf{Method} & & \textbf{\begin{tabular}[c]{c}GMACs\\$(\downarrow)$\end{tabular}}  & \textbf{\begin{tabular}[c]{c}Control\\Params.$(\downarrow)$\end{tabular}} & & \textbf{\begin{tabular}[c]{c}HV\\$(\uparrow)$\end{tabular}}\\
\cline{1-1}\cline{3-4}\cline{6-6}
PHN & & 21.02 & 21.04M & & 02.36 \\
PHN-BN & & 21.02 & 02.23M & & 11.72 \\
\hline
CDMA$^\dagger$ & & 25.98 & 00.03M & & 09.53\\
CDMA & & 29.04 & 00.03M & & 12.42 \\
\hline
MTL-Static  & & 22.66 & - & & 35.64 \\
\rowcolor{eq-fig-tab-color!10}
\altar$^\dagger$ & & 10.00 & 00.20M & & \textbf{34.91}\\
\rowcolor{eq-fig-tab-color!10}
\altar & & 20.00 & 00.20M && 34.89\\ 
\bottomrule
\end{tabular}}
\label{tab:nyud_perf}
\end{table}
\begin{table}[t]
\caption{\textbf{Evaluation on PascalContext}. Reference point for HV: $[3, 3, 3, 3, 3]$. $\dagger$ indicates smaller compute cost (GMACs).}
\vskip0.5em
\centering
\resizebox{\columnwidth}{!}{
\begin{tabular}{lccccc}
\toprule
\textbf{Method} & & \textbf{\begin{tabular}[c]{c}GMACs\\$(\downarrow)$\end{tabular}}  & \textbf{\begin{tabular}[c]{c}Control\\Params.$(\downarrow)$\end{tabular}} & & \textbf{\begin{tabular}[c]{c}HV\\$(\uparrow)$\end{tabular}}\\
\cline{1-1}\cline{3-4}\cline{6-6}
PHN & & 06.28 & 21.50M & & 42.61 \\
PHN-BN & & 06.28 & 03.63M & & 72.27 \\
\hline
CDMA$^\dagger$ & & 06.81 & 15.32M & & 73.20 \\
CDMA& & 07.21 & 15.32M & & 75.52 \\
\hline
MTL-Static  & & 09.43 & - & & 242.79  \\
\rowcolor{eq-fig-tab-color!10}
\altar$^\dagger$ & & 04.00 & 0.021M & & 166.75\\ 
\rowcolor{eq-fig-tab-color!10}
\altar & & 09.00 & 0.021M & & \textbf{166.97} \\ 
\bottomrule
\end{tabular}}
\label{tab:pas_perf}
\end{table}
\subsection{Comparison with Baselines} 
We present the overall multi-task performance of sampling sub-architectures with three different distributions NYUD-v2 (3 tasks), Pascal-Context (5 tasks) and CIFAR-100 (20 tasks) in Tab. \ref{tab:nyud_perf}-\ref{tab:cifar_perf}. Here we compare with state-of-the-art methods CDMA (allows compute budget controllability along with task controllability), PHN and PHN-BN (allows only task controllability). Following CDMA \cite{raychaudhuri2022controllable}, we report HV (reference point set according to CDMA) across 20 task preferences. Similar to CDMA, we analyze \altar under both small and large extremes of compute budgets (indicated using GMACs). We also account for the external networks used to sample sub-architectures under `Control Parameters'. For PHN and CDMA, this is the cost of the hypernetworks, while for \altar, the cost pertains to the MTL accuracy predictor $\mathcal{R}$.
Note that results for both CDMA and \altar are computed from one model while varying compute and task preference constraints. 
\begin{table}[t]
\caption{\textbf{Evaluation on CIFAR100-MTL}. Reference point for HV: $[1, 1, \cdots, 1]$. $\dagger$ indicates smaller compute cost (GMACs)} 
\vskip0.5em
\centering
\resizebox{\columnwidth}{!}{
\begin{tabular}{lccccc}
\toprule
\textbf{Method} & & \textbf{\begin{tabular}[c]{c}GMACs\\$(\downarrow)$\end{tabular}}  & \textbf{\begin{tabular}[c]{c}Control\\Params.$(\downarrow)$\end{tabular}} & & \textbf{\begin{tabular}[c]{c}HV\\$(\uparrow)$\end{tabular}}  \\
\cline{1-1}\cline{3-4}\cline{6-6}
PHN & & 073.13 &11.03M & & 0.002 \\
PHN-BN & & 073.13 &00.31M & & 0.007 \\
\hline
CDMA$^\dagger$ & & 129.23 &03.10M & & 0.009 \\
CDMA & &174.36 & 03.10M & &\textbf{0.010} \\
\hline
MTL-Static  & & 072.92 & - & & 0.0089  \\
\rowcolor{eq-fig-tab-color!10}
\altar$^\dagger$ & & 038.40 & 00.02M & & 0.0026 \\ 
\rowcolor{eq-fig-tab-color!10}
\altar & & 072.92 & 00.02M & & 0.0082  \\ 
\bottomrule
\end{tabular}}
\label{tab:cifar_perf}
\end{table}
\begin{figure}[!t] 
\vspace{-1em}
\begin{subfigure}[b]{0.33\linewidth}
    \centering
    \includegraphics[width=\linewidth]{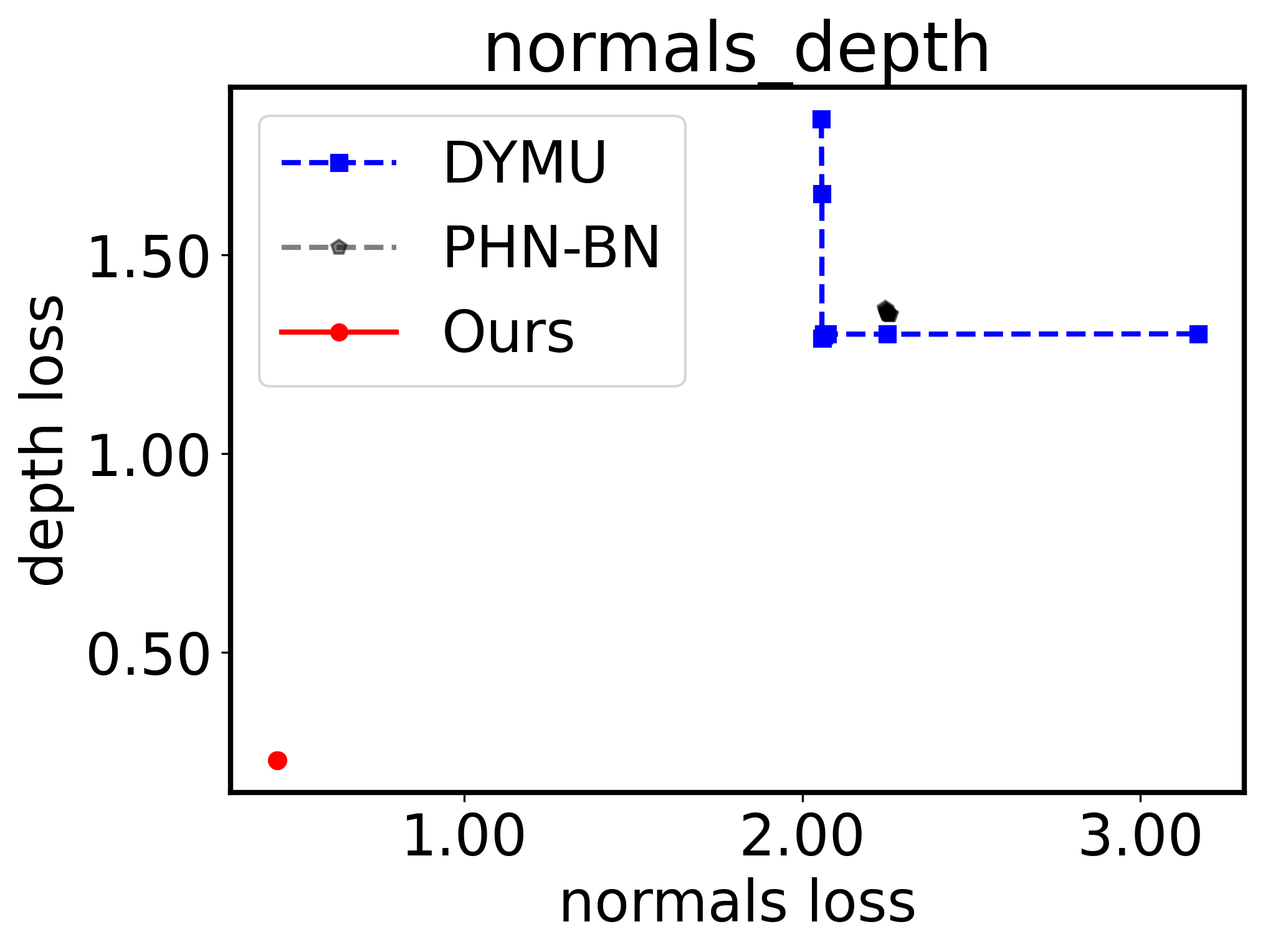} 
    \label{fig:nyud_pareto_1} 
  \end{subfigure}
  \begin{subfigure}[b]{0.33\linewidth}
    \centering
    \includegraphics[width=\linewidth]{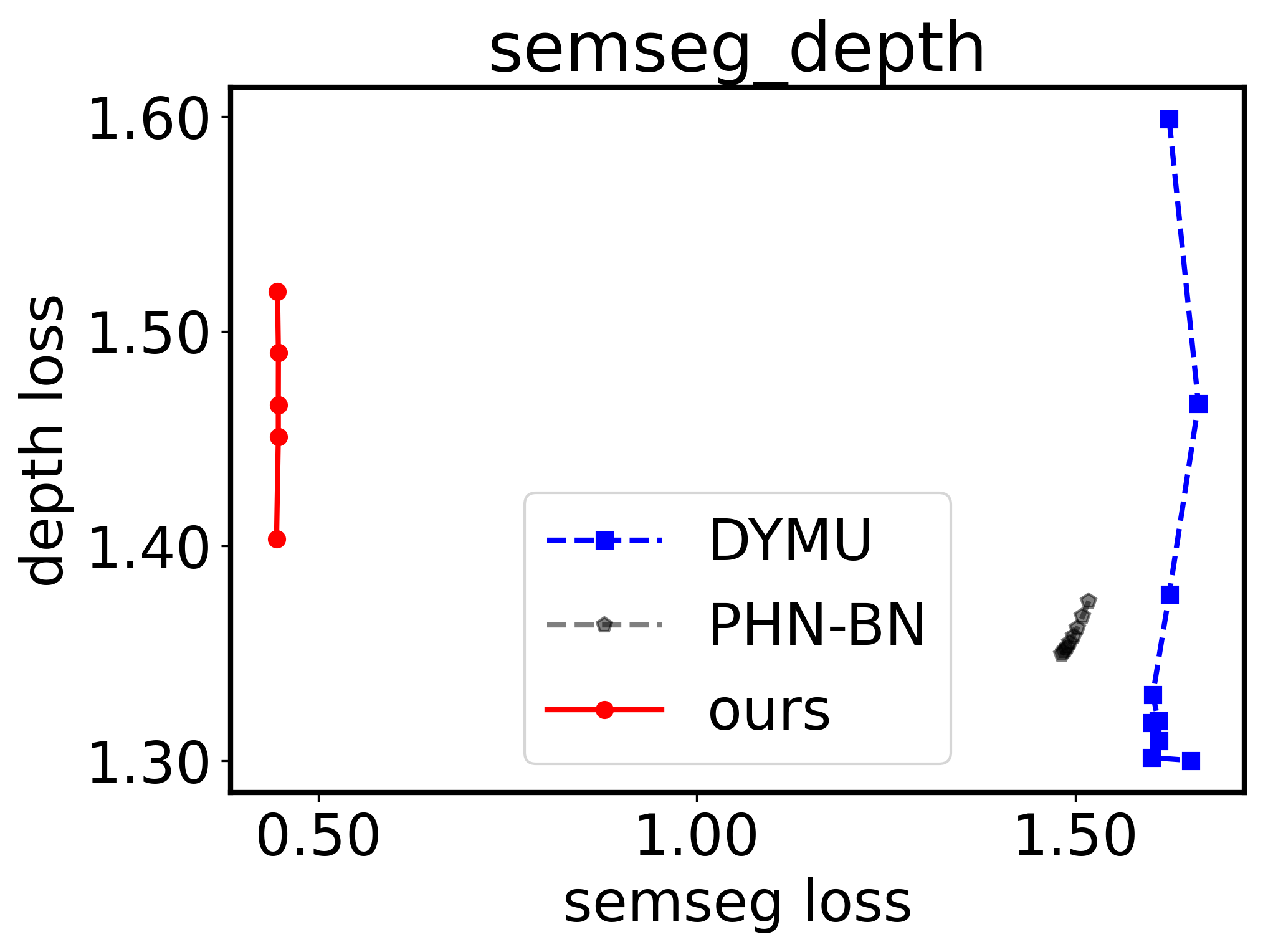} 
    \label{fig:nyud_pareto_2} 
  \end{subfigure} 
  \begin{subfigure}[b]{0.33\linewidth}
    \centering
    \includegraphics[width=\linewidth]{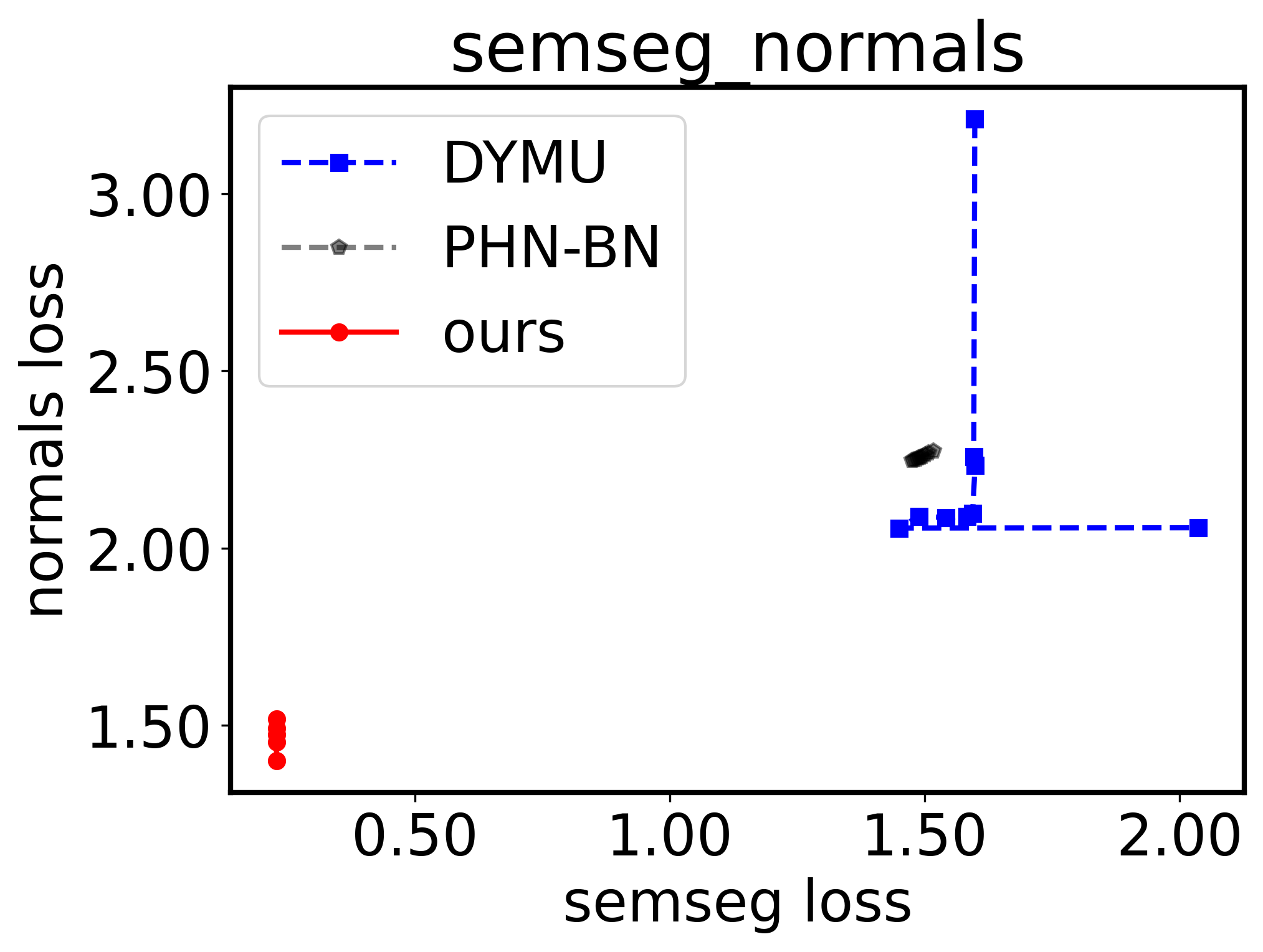} 
    \label{fig:nyud_pareto_3} 
  \end{subfigure}
  \vspace{-2ex}
  \caption{\textbf{Loss curves comparison on NYUD-v2 (3 tasks).} Due to strategy to translate task preferences to task decoders, task losses remain low even if the preference decreases. This supports our training paradigm to improve the performance of the smaller architectures, not done in CDMA and PHN.}
  \label{fig:loss_curve} 
\end{figure}

\altar demonstrates high HV values compared to the existing methods, especially in NYUD-v2 and Pascal-Context. The high HV along with lower GMACs supports our claim that the proposed shared encoder and individual decoders are capable of solving multi-task problems while handling task trade-offs. For example, \altar shows a higher HV by $\sim$ 33.5\% in the NYUD-v2 dataset and $\sim$ 55\% in Pascal-Context dataset, while using comparatively much less compute in both the cases. We believe this strong performance is due to our strategy to incorporate task decoders in the search space which is ignored by CDMA. For the classification task of CIFAR-100, \altar achieves similar HV with lesser compute cost. We believe that the impact of our task preference translation as per Eq. \ref{eq:dec_rule} is smaller for CIFAR-100 given that the decoder of the classifier (fully connected layers) is computationally lighter than dense prediction decoders. Finally, our strategy to translate the task preferences to the decoders allows us to maintain the HV even with highly restricted compute cost. This is highly advantageous compared to PHN and CDMA, where the change in control parameters or the compute cost deteriorates the HV.
\begin{table}[t]
\caption{\textbf{Ablation Study on NYUD-v2.} \T{1}: Semantic Seg., \T{2}: Human-Parts Seg., \T{3}: Depth Est., Here, the colored row indicates performance of largest model for reference. Bottom two rows indicate the impact of loss $\Lkd$ on the performance of smallest possible models for our method. HV is computed by keeping the budget indicated under GMACs for all the cases.}
\vskip0.4em
\resizebox{\columnwidth}{!}{%
\begin{tabular}{cccccccccc}
\toprule
\multicolumn{2}{c}{\textbf{Learning Loss}} && \multirow{2}{*}{\textbf{GMACs}} && \multicolumn{3}{c}{\textbf{Task Performance}} && \textbf{Control} \\ \cline{1-2}\cline{6-8}\cline{10-10}
$\Lco$ & $\Lkd$ & & & & \T{1} ($\uparrow$) & \T{2} ($\downarrow$) & \T{3} ($\downarrow$) & & \textbf{HV} ($\uparrow$) \\\hline
\rowcolor{red!10}
-- & -- && 22.66 &&  36.03 & 26.82 & 0.63 & & 34.86 \\
\cline{1-10}
\ccheck & \ccross && && 30.35 & 28.86 & 0.70 && 32.87 \\
\ccheck & \ccheck && \multirow{-2}{*}{08.39} && 32.38 & 28.52 & 0.67 && 33.81 \\
\bottomrule
\end{tabular}%
}
\label{tab:ablation_table}
\end{table}
\begin{figure}[!t] 
\vspace{-0.5em}
\begin{subfigure}[b]{0.5\linewidth}
    \centering
    \includegraphics[width=0.9\linewidth]{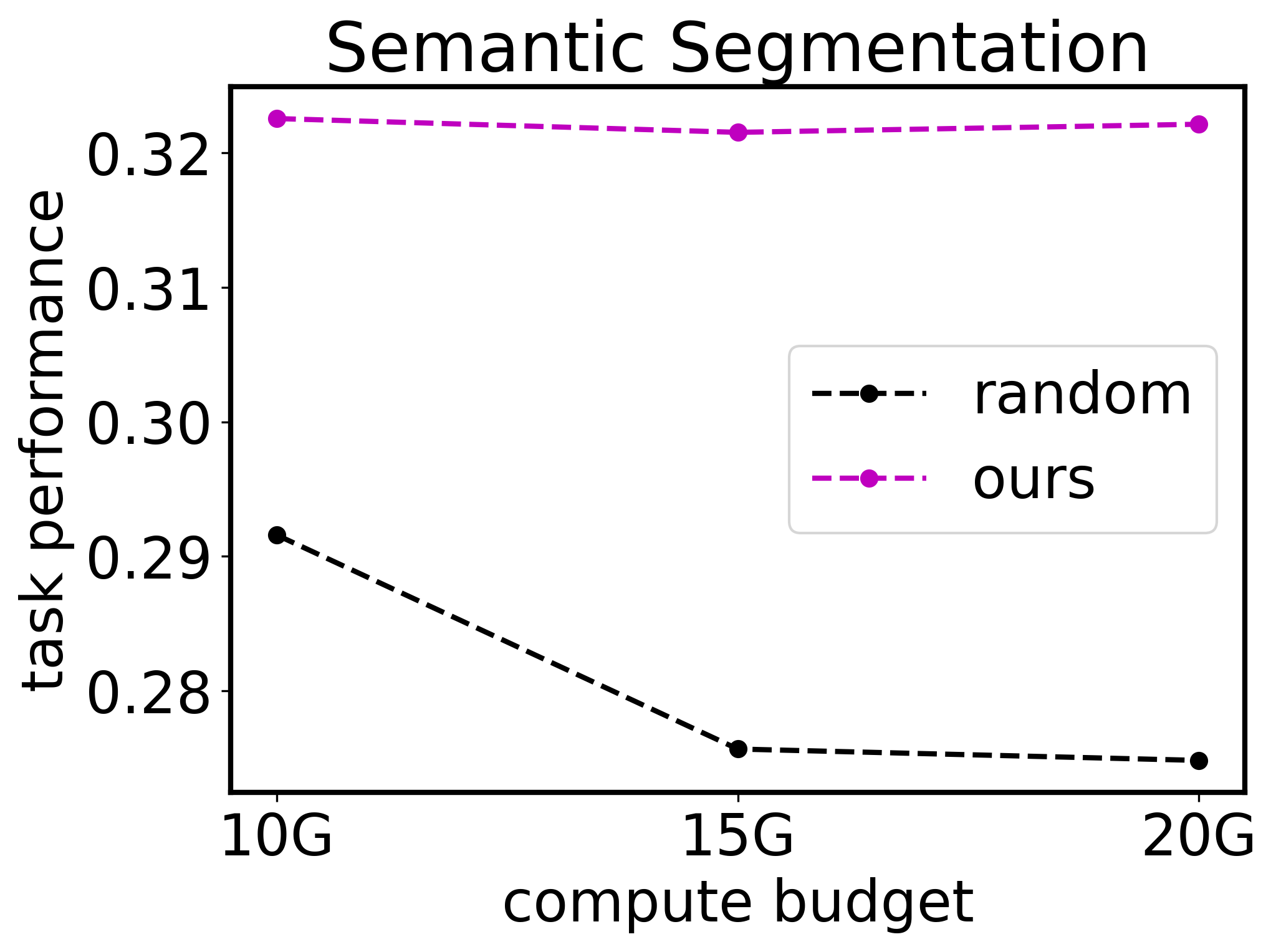} 
    \label{fig:nyud_search_1} 
  \end{subfigure}
  \begin{subfigure}[b]{0.5\linewidth}
    \centering
    \includegraphics[width=0.9\linewidth]{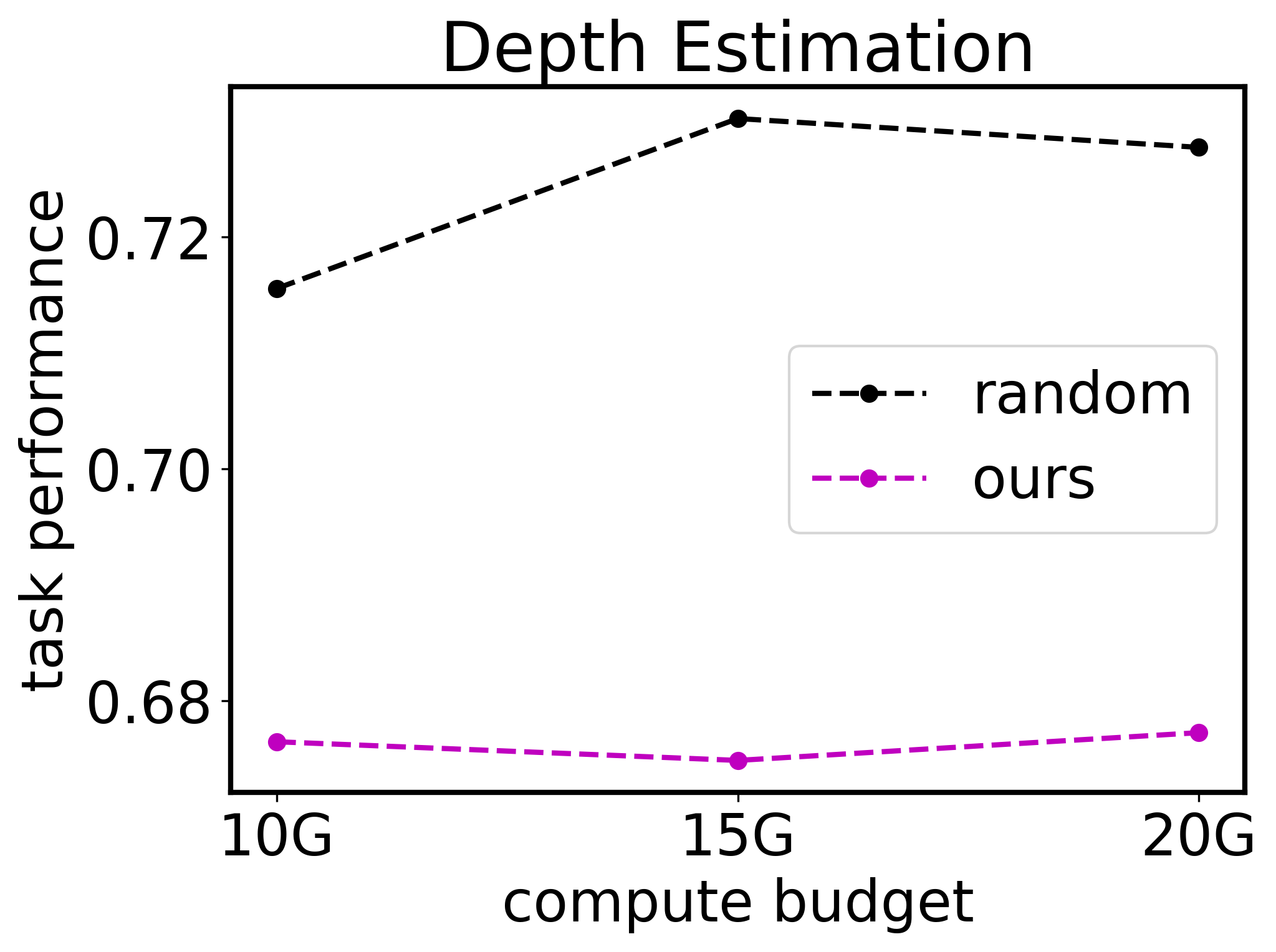} 
    \label{fig:nyud_search_2} 
  \end{subfigure} 
  \caption{\textbf{Search algorithm comparison on NYUD-v2}. Sematic seg.: \textit{higher} and Depth est.: \textit{lower} is better. Our proposed search algorithm provides better architectures with negligible search cost.}
  \label{fig:search_comparison} 
\end{figure}

As the HV is a function of the loss values computed over changing task preferences, we visualize the task preference - task loss trade-off curves in Fig. \ref{fig:loss_curve}. Due to our strategy to translate task preferences to task decoders, the task losses remain low even if the task preference decreases. This is because lower width SubNets (created due to lower task preferences) are enhanced to perform closely to the SuperNet. This further attests to our training paradigm to improve the performance of the smaller MTL architectures, a feature absent in CDMA and PHN. 
\begin{figure}[!t] 
\begin{subfigure}[!b]{0.32\linewidth}
    \centering
    \includegraphics[width=0.945\linewidth]{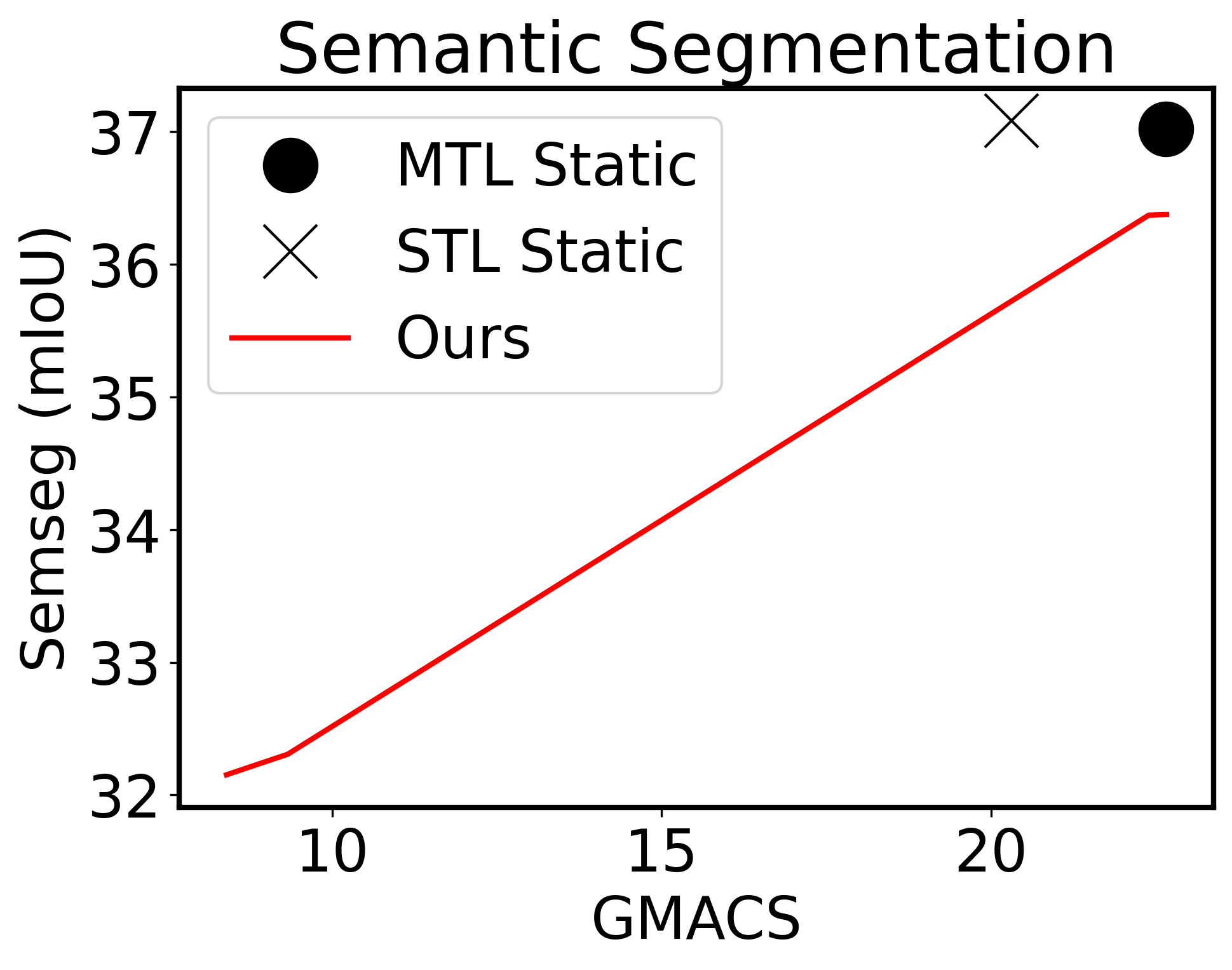} 
    \label{fig:nyud_task_a} 
  \end{subfigure}
  \begin{subfigure}[!b]{0.32\linewidth}
    \centering
    \includegraphics[width=0.955\linewidth]{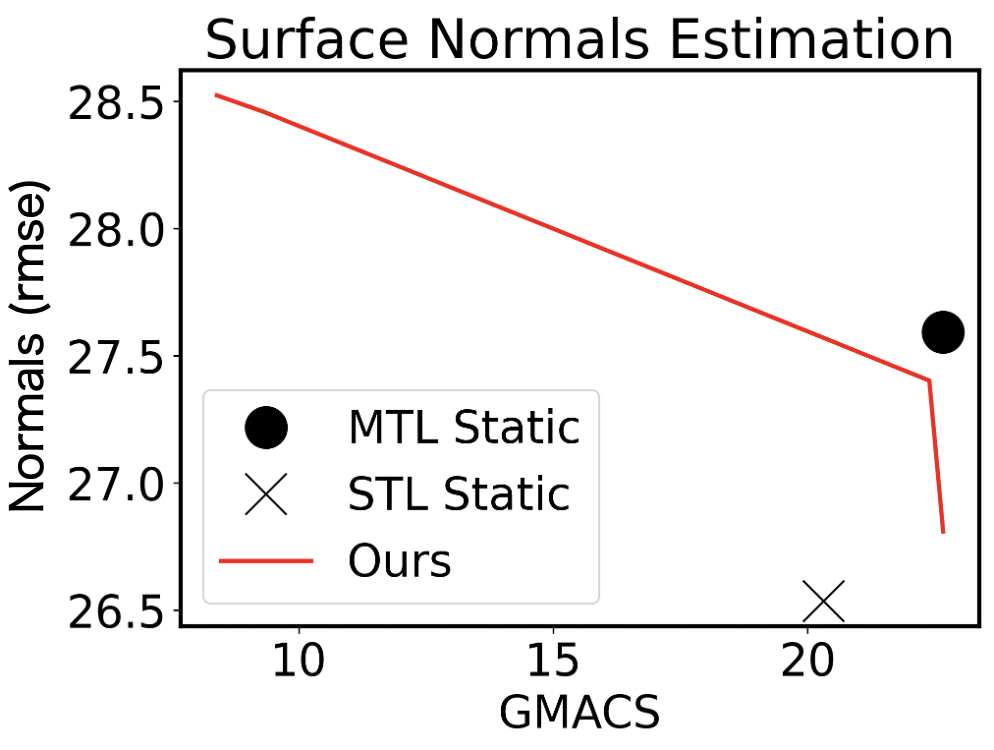} 
    \label{fig:nyud_task_b} 
  \end{subfigure}
  \begin{subfigure}[!b]{0.32\linewidth}
    \centering
    \includegraphics[width=0.955\linewidth]{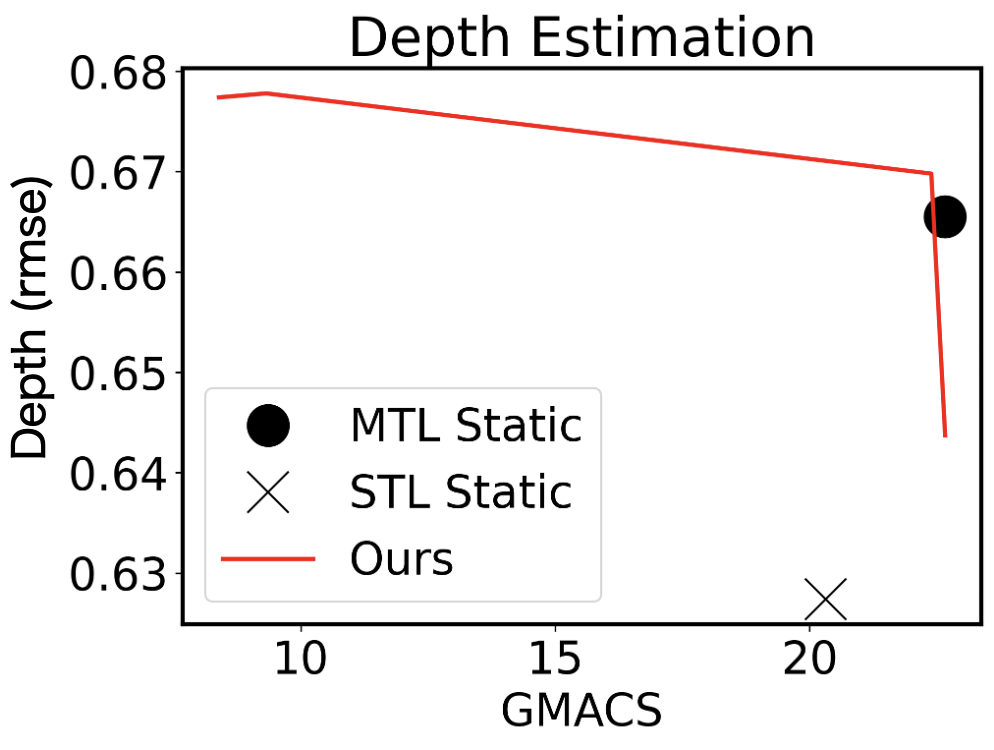} 
    \label{fig:nyud_task_c} 
  \end{subfigure}
  \vspace*{0.25em}
  \caption{\textbf{Saving design costs.} NYUD-v2 (3 tasks) with ResNet34 backbone. Our training method allows to sample multiple high performing MTL architectures based on the user compute cost after training only once, in contrast to static MTL/STL architectures where the training has to be repeated for changing constraints. }
  \label{fig:nyud_task_performance} 
\end{figure}
\subsection{Ablation Study} 
\begin{figure}[!t]
    \centering
    \includegraphics[width=0.9\columnwidth]{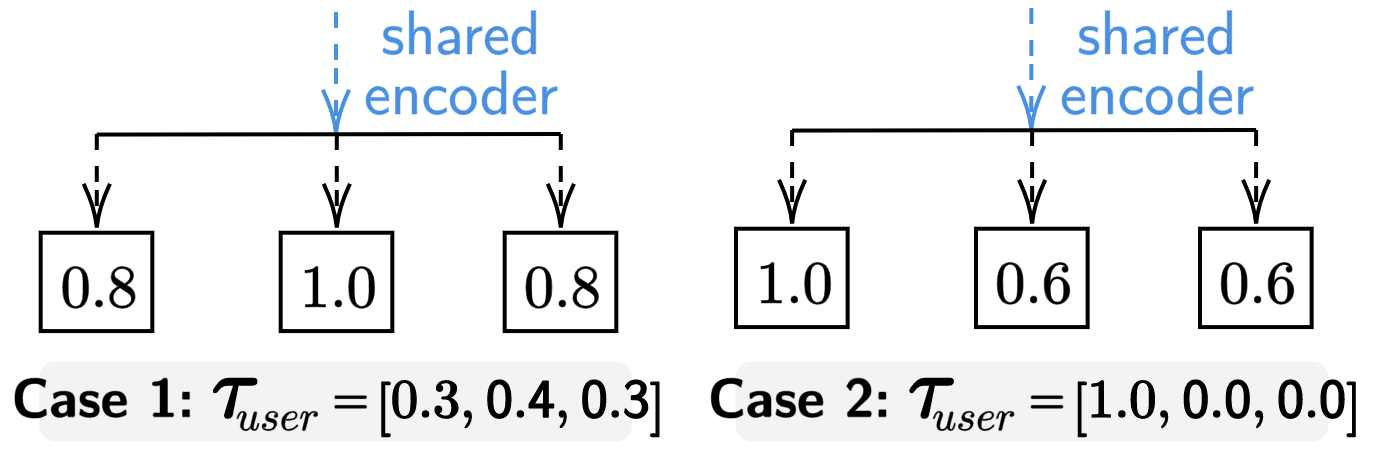}
    \caption{\textbf{Illustration of our task preference translation.} We visualize the width ratios computed for the task decoders for NYUD-v2 (3 tasks). Note that we normalize the task preference using max-min normalization before computing the decoder widths.}

    \label{fig:supp_decoder_task_pref}
\end{figure}

\paragraph{Effectiveness of CI-KD loss and shared encoder size.} In Tab. \ref{tab:ablation_table}, we analyze the importance of proposed loss functions in our SuperNet training using two different backbones. It can be observed that the CI-KD loss $\Lkd$ pushes the performance of the smallest network towards the parent network for both the cases. For example, for the  ResNet34 backbone SuperNet, CI-KD loss increases the semantic segmentation by almost 2 points whereas the HV increases by 1 point for the smallest architecture with 37\% less compute cost. 
\paragraph{Effectiveness of our search strategy.} Our search algorithm is aimed to find the best shared encoder configuration that supports the sampled task decoders based on Eq. \ref{eq:dec_rule}. To show its efficacy, we compare it with the strategy of choosing the width ratios of the encoder randomly in Fig. \ref{fig:search_comparison} for the NYUD-v2 dataset. We train a MTL SuperNet with the training set $\D_{tr}$. The `random' search algorithm then samples width ratios from the available width list $\WList$ and chooses the encoder configuration satisfying the user's compute budget. Our proposed search, on the other hand, uses the accuracy predictor $\mathcal{R}$ created with the validation set $\D_{val}$ and chooses an optimal encoder configuration that supports the decoder for best task performance. Note that the time for our search strategy is negligible compared to training a sub-architecture from scratch. Further, computation overhead due to our search algorithm is only needed when the user preferences change and does not influence the final inference time.   
%
\begin{figure}[!t] 
\begin{subfigure}[b]{0.33\linewidth}
    \centering
    \includegraphics[width=\linewidth]{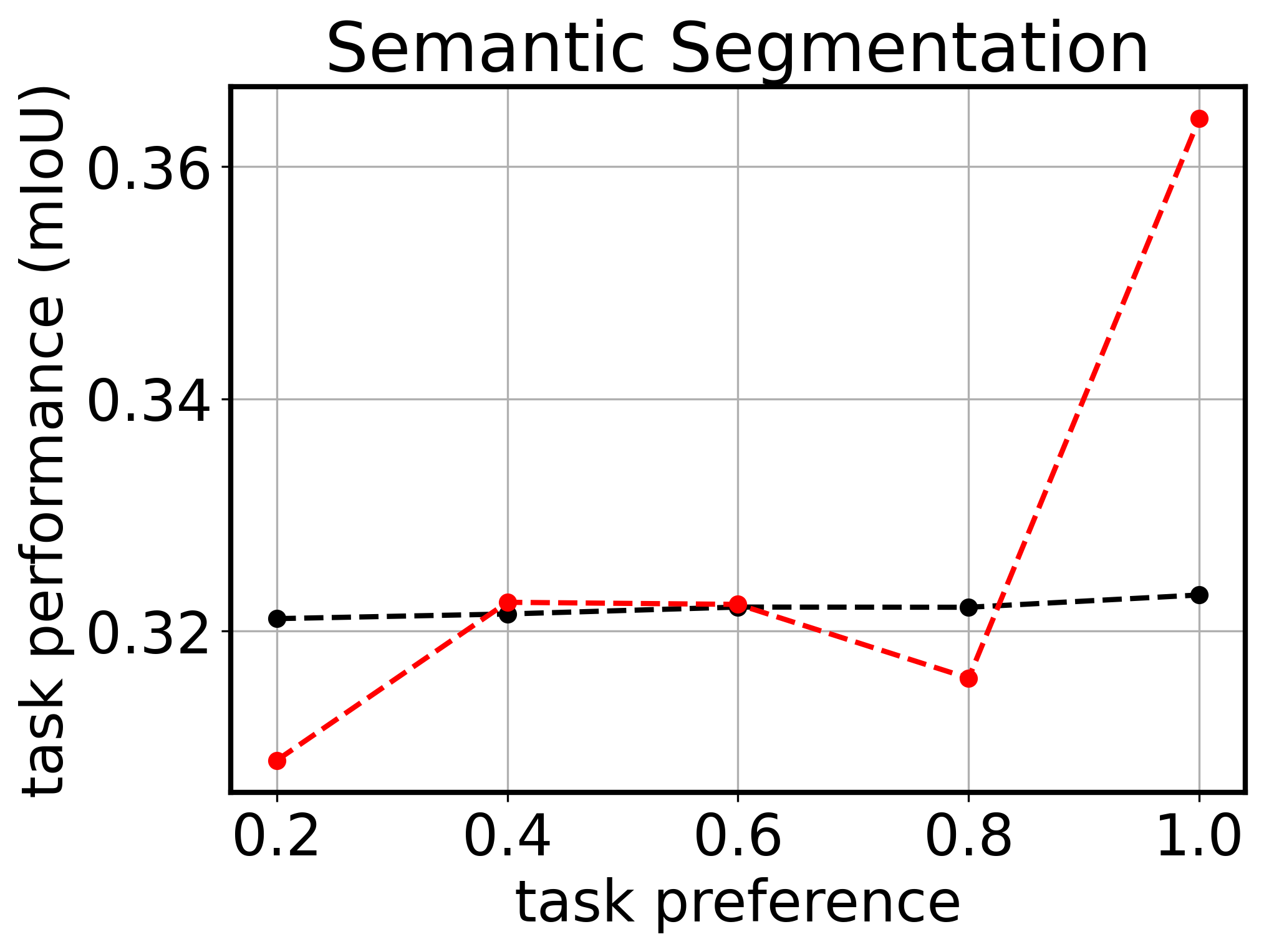} 
    \label{fig:nyud_marginal_a} 
  \end{subfigure}
  \hfill
  \begin{subfigure}[b]{0.33\linewidth}
    \centering
    \includegraphics[width=\linewidth]{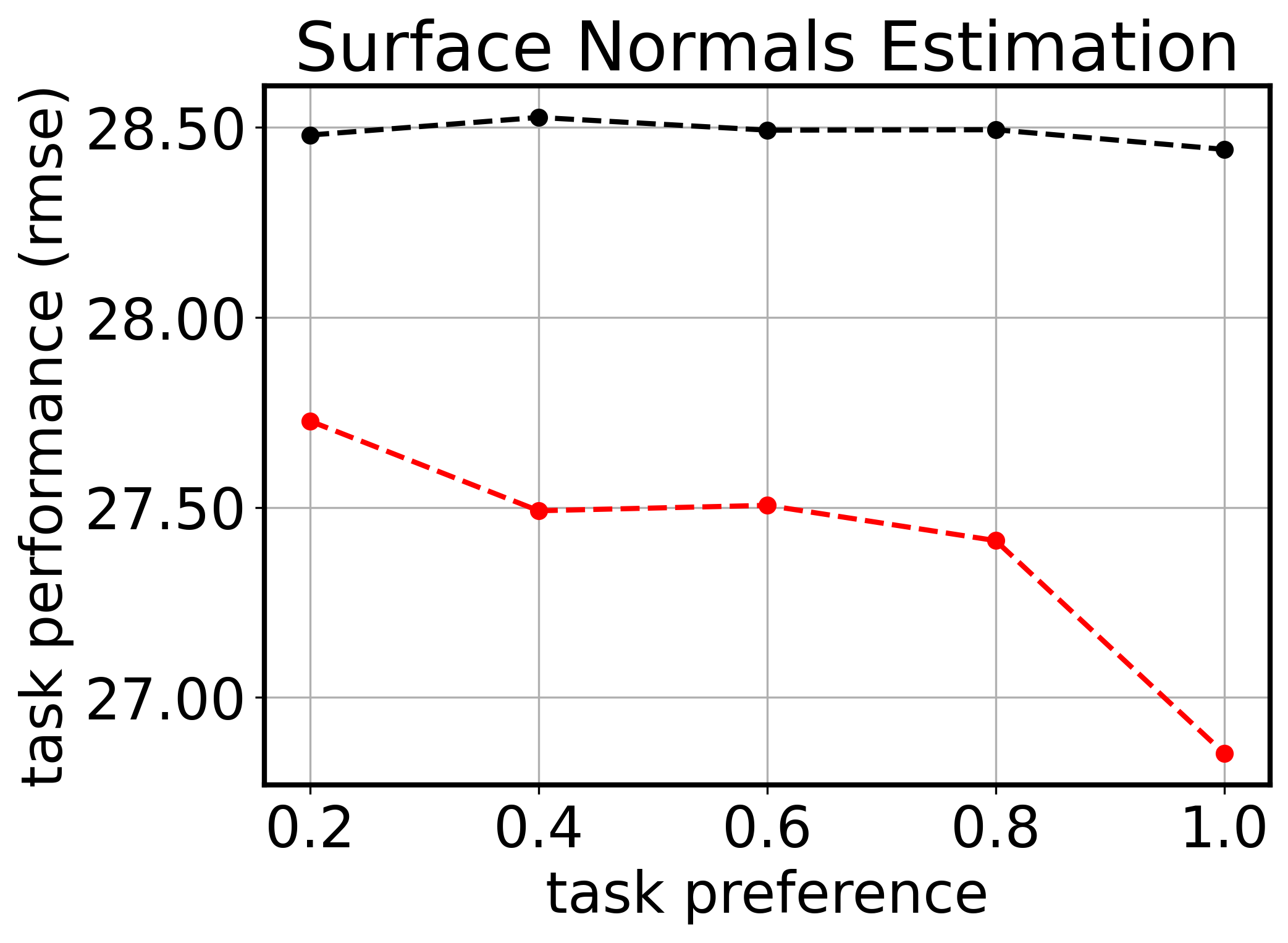} 
    \label{fig:nyud_marginal_b} 
  \end{subfigure}
  \hfill
  \begin{subfigure}[b]{0.33\linewidth}
    \centering
    \includegraphics[width=\linewidth]{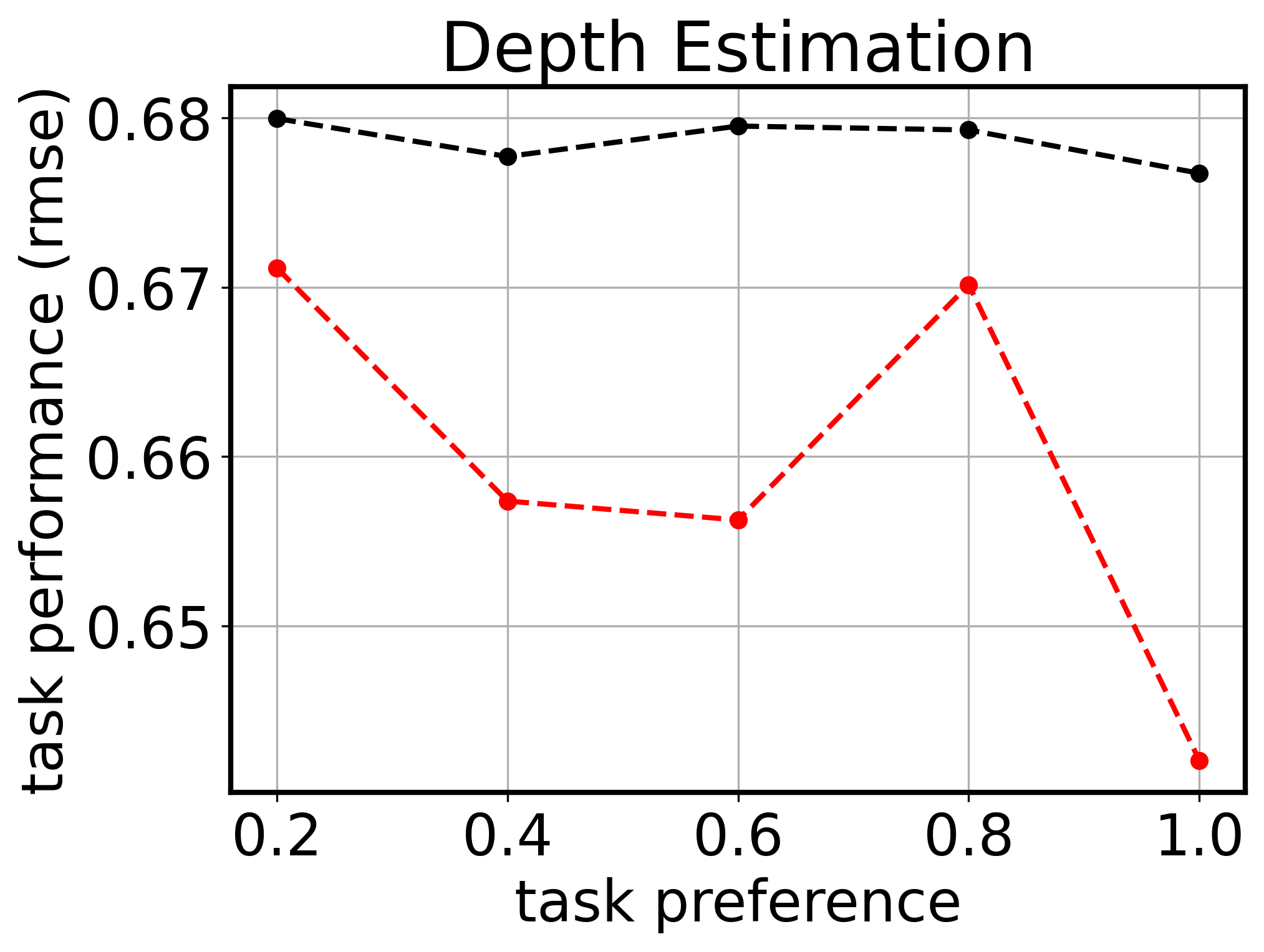} 
    \label{fig:nyud_marginal_c} 
  \end{subfigure}
  \caption{\textbf{Marginal evaluation on NYU-v2.} Our proposed method to translate task preference to task decoders is effective in two encoder configurations (\textcolor{red}{largest}, \textbf{smallest}). Different tasks have different response to encoder configuration which shows the necessity of our search algorithm to find an optimal encoder configuration.}
  \label{fig:nyud_marginal} 
\end{figure}
\paragraph{Marginal task evaluation.} In Fig. \ref{fig:nyud_marginal}, we visualize the task performance for \altar when the task preference changes, by marginalizing over preference values of the other tasks. This is done for two extreme cases of the encoder configuration: by setting the encoder widths to the largest and smallest size, respectively. Clearly, an increase in task preference leads to an increase in the task performance. This is a direct result of our proposed method to set the decoder width ratios according to Eq. \ref{eq:dec_rule}. Further, it also shows that different tasks respond differently to the encoder configurations. For example, the task of semantic segmentation performs almost similarly for the task preference range [0.2, 0.8] for the both configurations, whereas the other tasks clearly present a bigger difference. This shows the necessity of our search algorithm to find a better encoder configuration. Fig. \ref{fig:nyud_marginal} also shows the accuracy-compute trade-off available per task.
\paragraph{Savings on design costs.} Along with the aforementioned advantages of task performance tradeoff, our training method in Sec. \ref{sec:method} allows to sample multiple high performing MTL architectures while training only once. This saves expensive training costs and allows multiple deployments from one model. We demonstrate this feature of \altar in Fig. \ref{fig:nyud_task_performance} for NYUD-v2 (3 task) dataset for ResNet34 backbone. Here, the static MTL and STL architectures can only be used for a fixed compute budget. In contrast, \altar can operate with variable compute costs and gracefully degrade in performance with decreasing budget. Note that one can further enhance the MTL performance of our parent architecture using state-of-the-art methods summarized in \cite{vandenhende2021multi, zhang2021survey}. 
\paragraph{Qualitative examples.}
We visualize some examples of the task decoder widths computed using the task preferences from \texttt{Eq.(4)} in Fig. \ref{fig:supp_decoder_task_pref}. Each block represents the task decoder with its selected width ratio for the given $\TList_{user}$. The design choice of $\WList$ for NYUD-v2 is $[0.6, 0.7, \cdots, 1.0]$. 

\vspace*{-\baselineskip}
\section{Conclusions}
We tackle the problem of multi-task architecture deployment based on user joint preferences (task performance \textit{and} available compute). The core challenge is to train the multi-task architecture \textit{once} but permit instant customization of the network for diverse preferences. We propose to train non-uniformly slimmable MTL architectures parameterized by layer wise width configurations, resulting in a large space of MTL architectures. At inference, we use an evolution-based search algorithm to sample precise MTL sub-architectures based on the joint preferences. In contrast to prior controllable MTL methods solely focused on encoders and providing poor task scalability, our method uses a shared encoder and the task decoders as part of search space. We show the effectiveness of our method on various multi-task settings, providing large search space for sampling efficient MTL models covering a wide range of MTL preferences.

\paragraph{Limitations and Future Works.} Our work has the following limitations that presents new opportunities for interesting future works. ECMT has only been explored for CNN based architectures and should be investigated for transformer based models. Further, the task relativity should be considered when extracting a sub-network as it is considered an important factor in MTL.   
\begin{spacing}{0.95}
{\small\paragraph{Acknowledgements.} This work was a part of Abhishek Aich’s internship at NEC Labs America. This work was also partially supported by NSF grant 1724341. We would like to thank the authors of \cite{cai2020once} for providing the implementations related to accuracy predictor used in their paper.}
\end{spacing}

\FloatBarrier
{\small
\bibliographystyle{unsrtnat}
\bibliography{egbib}
}

\end{document}